\documentclass[12pt]{article}
\usepackage{spconf,amsmath,graphicx}
\usepackage{eso-pic}
\usepackage{amssymb}
\usepackage{epstopdf}
\usepackage{relsize}

\usepackage{tikz}
\usetikzlibrary{arrows,shapes}
\usetikzlibrary{positioning}

\usepackage{pgfplots}
\usepgfplotslibrary{fillbetween}
\pgfplotsset{compat=newest}

\newlength\figureheight
\newlength\figurewidth
\def \landmark {\boldsymbol{\kappa} }

\title{A weighting strategy for Active Shape Models}

\name{Alma Eguizabal and Peter J. Schreier
\thanks{Copyright 2017 IEEE. Published in the IEEE 2017 International Conference on Image Processing (ICIP 2017), scheduled for 17-20 September 2017 in Beijing, China. Personal use of this material is permitted. However, permission to reprint/republish this material for advertising or promotional purposes or for creating new collective works for resale or redistribution to servers or lists, or to reuse any copyrighted component of this work in other works, must be obtained from the IEEE. Contact: Manager, Copyrights and Permissions / IEEE Service Center / 445 Hoes Lane / P.O. Box 1331 / Piscataway, NJ 08855-1331, USA. Telephone: + Intl. 908-562-3966.}}

\address{Signal and System Theory Group, University of Paderborn, Germany
\\ e-mail: \{alma.eguizabal, peter.schreier\}@sst.upb.de}

\begin{document}
\ninept
\maketitle
\begin{abstract}
Active Shape Models (ASM) are an iterative segmentation technique to find a landmark-based contour of an object. In each iteration, a least-squares fit of a plausible shape to some detected target landmarks is determined. Finding these targets is a critical step: some landmarks are more reliably detected than others, and some landmarks may not be within the field of view of their detectors. To add robustness while preserving simplicity at the same time, a generalized least-squares approach can be used, where a weighting matrix incorporates reliability information about the landmarks. We propose a strategy to choose this matrix, based on the covariance of empirically determined residuals of the fit. We perform a further step to determine whether the target landmarks are within the range of their detectors. We evaluate our strategy on fluoroscopic X-ray images to segment the femur. We show that our technique outperforms the standard ASM as well as other more heuristic weighted least-squares strategies.
\end{abstract}

\begin{keywords}
Active Shape Models, Generalized Least-Squares, fluoroscopic X-rays.
\end{keywords}

\section{Introduction}
\label{sec:intro}
Active Shape Models (ASM) \cite{Cootes96} are a segmentation technique widely used in computer vision and image processing. They find an object contour parameterized by a set of landmarks using an iterative least-squares fit that minimizes the distance between a plausible object shape and detected target landmarks. ASM adapt to many kinds of shapes and imaging modalities. However, they do require images with reasonably good contrast in order to find the object contours. An example of a challenging scenario is intraoperative fluoroscopic X-ray imaging, which produces low-quality images due to the low X-ray dose. More robust alternatives to ASM exist (e.g., \cite{Lindner13}), but they are typically more computationally expensive. In this paper, we show that we can keep the simplicity of ASM and still improve robustness by replacing the least-squares procedure with a generalized least-squares (GLS) approach. The idea of GLS is to weight individual landmarks according to their reliability: reliably identified landmarks should be trusted more than less reliable landmarks. This raises the question of how to identify the right weighting strategy.

This question has already been addressed before: in \cite{Hill96}, where directional regularization is proposed; in \cite{Rogers02}, which suggests to use robust parameter estimation; in \cite{Zhao04} and \cite{Ruiz16}, where the weights change in every iteration depending on a score of the target detectors; in \cite{Lekadir07}, which proposes a pose-invariant metric; and in \cite{Yang14}, which measures the reliability of the landmarks based on models of local appearance. Many of these weighting strategies are heuristic and also prone to over-fitting. Some of them also add computational complexity to the ASM algorithm. 

Our contribution is a strategy for choosing the weights, which has a theoretical justification and still keeps the simplicity of the ASM algorithm. Our proposed strategy measures the reliability of the target landmarks based on the covariance of the residuals of the fit obtained from training data. Additionally, we perform a test to determine whether a landmark is valid, i.e., whether it is within the field of view of its target detector, and we incorporate this test into the weighting matrix. 

Our program for this paper is as follows. Section 2 summarizes the ASM algorithm and introduces the GLS approach. In Section 3 we discuss our approach for choosing the weights in the GLS optimization. We also introduce our test for determining whether a landmark is valid. In Section 4, we evaluate the performance of our technique based on a leave-one-out test. Finally, conclusions are drawn in Section 5.

\section{Problem formulation}
\label{sec:formul}

We use ASM to segment an object contour in a 2D image. This segmentation places a set of $N$ landmarks on the contour of the object of interest. The variability of the position of these landmarks for a particular object is modeled in terms of pose and shape: pose refers to any rotation, scale and translation; shape refers to the variability remaining after transforming the landmarks into coordinates with common pose. We use $M$ training images with manually placed landmarks to learn the shape variability. We model the landmarks in a complex space  $ \mathbb{C}$, where the pose parameters are defined by a complex affine transformation: the landmarks in the $m$th training image are complex vectors $ \landmark_m \in  \mathbb{C}^{N \times 1}$, whose real and imaginary parts correspond to the coordinates of the two-dimensional Euclidean space. 

To model the shape variability, ASM use a Point Distribution Model (PDM). To build the PDM, we perform a Procrustes alignment \cite{BOOKSTEIN97} to find the coordinate space of the common pose. The pose-aligned vectors are $ \mathbf{k}_m = r_m\landmark_m+\mathbf{1}t_m $, where $\{r_m,t_m\} \in \mathbb{C}$ are the pose parameters. Let us define an invertible function $ V : \mathbb{C}^{N \times 1}\rightarrow \mathbb{R}^{2N \times 1} $ as $V(\mathbf{k}) = [\operatorname{Re}(\mathbf{k})^T,  \operatorname{Im}(\mathbf{k})^T]^T$, which provides the corresponding real description. We also define the vector $\mathbf{x}_m = V(\mathbf{k}_m)\in \mathbb{R}^{2N\times 1}$ and the matrix $ \mathbf{X} = [\mathbf{x}_1,\mathbf{x}_2,\dots,\mathbf{x}_M] \in \mathbb{R}^{2N\times M}$ that contains the $M$ aligned training vectors. 

We then compute the sample mean of these vectors as  $\hat{\boldsymbol{\mu}}_\mathbf{x} = \frac{1}{M}\sum_{m=1}^{M}\mathbf{x}_m$ as well as the sample covariance matrix $\hat{\mathbf{S}}_{\mathbf{x}} = \frac{1}{M-1}(\mathbf{X}\mathbf{X}^T-\hat{\boldsymbol{\mu}}_\mathbf{x}\hat{\boldsymbol{\mu}}_\mathbf{x}^T)$. After decomposing $\hat{\mathbf{S}}_{\mathbf{x}}$ into its eigenvectors $\mathbf{U}$ and eigenvalues $\boldsymbol{\Lambda}$, i.e. $\hat{\mathbf{S}}_{\mathbf{x}} = \mathbf{U}\boldsymbol{\Lambda}\mathbf{U}^T$, we keep the $p$ most significant eigenvectors in a matrix $\mathbf{P} = [\mathbf{u}_1, \mathbf{u}_2, \ldots, \mathbf{u}_p] \in \mathbb{R}^{2N\times{p}}$. Consequently, $\mathbf{x}_m$ can be linearly approximated as $\mathbf{x}_m \approx \hat{\boldsymbol{\mu}}_{\mathbf{x}} + \mathbf{Pb}_m$, and  therefore any vector of landmarks can be approximated as
\begin{equation} \label{eq1} 
\landmark_m \approx \frac{1}{r_m}\Big(V^{-1}(\hat{\boldsymbol{\mu}}_{\mathbf{x}}  +\mathbf{Pb}_m)-\mathbf{1}t_m \Big)\;,
\end{equation}
where $\mathbf{b}_m \in \mathbb{R}^{p\times1}$ is the shape parameter vector of $\landmark_m$, and  $\{r_m,t_m\}$ its pose parameters. 

We now briefly describe the iteration process of ASM. Let us assume that $\landmark^{(i)}$ is the resulting vector of landmarks after iteration $i$. Each iteration first examines local regions around each landmark in $\landmark^{(i-1)}$ in order to detect new target landmarks $\tilde{\landmark}^{(i)}$. However, $\tilde{\landmark}^{(i)}$ may not describe a plausible shape or an accurate pose. In order to find the closest plausible shape and pose, that is, the landmarks in vector $\landmark^{(i)}$, the following least-squares problem is solved: 
\begin{align}\label{min1}
\underset{r^{(i)}, t^{(i)},\mathbf{b}^{(i)}\in \mathbb{B}}{\min}
& ||\landmark^{(i)}-\tilde{\landmark}^{(i)}||^2 \;,
\end{align} 
where $\landmark^{(i)}=\frac{1}{r^{(i)}}[V^{-1}(\hat{\boldsymbol{\mu}}_{\mathbf{x}} + \mathbf{P}\mathbf{b}^{(i)}) -\mathbf{1}t^{(i)}]$, that is, $r^{(i)},t^{(i)}$ are the pose parameters and $\mathbf{b}^{(i)}$ is the shape parameter vector. To enforce shape plausibility,  $\mathbf{b}^{(i)}$ must be contained in $\mathbb{B}=\{\mathbf{b} \in \mathbb{R}^{p\times1} : \mathbf{b}^{T}\boldsymbol{\Lambda}_p^{-1}\mathbf{b}\leq \xi\}$, where $\boldsymbol{\Lambda}_p$ is a diagonal matrix with the $p$ most significant eigenvalues corresponding to the matrix $\mathbf{P}$, and $\xi$ is a suitable threshold. The expression in \eqref{min1} is an ordinary least-squares problem, where every component of the difference between the vectors $\landmark^{(i)}$ and $\tilde{\landmark}^{(i)}$ has the same impact on the minimization. These components, i.e. the landmark positions, can be weighted according to their reliability in a generalized least-squares (GLS) problem:
\begin{align}\label{minW}
\underset{r^{(i)}, t^{(i)},\mathbf{b}^{(i)}\in \mathbb{B}}{\min}
& (\landmark^{(i)}-\tilde{\landmark}^{(i)})^H\mathbf{W}(\landmark^{(i)}-\tilde{\landmark}^{(i)}) \; ,
\end{align} 
where $\mathbf{W}$ is the matrix of weights that controls the importance of individual landmarks in the optimization. We will consider both diagonal and nondiagonal weight matrices.

\section{Proposed Solution}
\label{sec:typestyle}
As described in Section 2, in each ASM iteration we need to find the new targets  $\tilde{\landmark}^{(i)}$. We determine these by means of a detector $\mathcal{T}(\landmark^{(i-1)}) = \tilde{\landmark}^{(i)}$, which explores the local regions around the previous landmarks $\landmark^{(i-1)}$. We assume that this detector searches for the best match of the gray-level profile around each landmark as measured by the Mahalanobis distance [1]. However, our technique can be generalized to other strategies as well. Section 3.1 and 3.2 describe how to choose the weighting matrix $\mathbf{W}$ based on the empirically determined residual errors. 

A complicating factor is that the true landmarks $\landmark^\star$ are not necessarily within the local regions explored by the detector $\mathcal{T}(\landmark^{(i-1)})$, i.e. they are not within its field of view. This is the case when the true landmarks are occluded or out of alignment (especially during the first iterations of the ASM algorithm). We therefore incorporate into our strategy a test of whether or not the true landmark is visible to the detector. If the test determines that a landmark is not visible, then this landmark is excluded by setting the corresponding entry in the weighting matrix $\mathbf{W}$ to zero. This is described in Section 3.3.

\subsection{GLS as a Maximum Likelihood (ML) problem}
\label{ssec:WLS}

Let us first assume the true landmarks $\landmark^\star$ are within the field of view of the detector $\mathcal{T}$ in iteration $i$, i.e. when determining $\tilde{\landmark}^{(i)}$. We define the residual error vector as $\boldsymbol{\epsilon}^{(i)}=\landmark^{(i)}-\tilde{\landmark}^{(i)}$.
We also assume that $\boldsymbol{\epsilon}^{(i)}$ is complex normal with zero mean and covariance matrix $\mathbf{R}$. Thus, the likelihood of the pose and shape parameters $\{r^{(i)}, t^{(i)},\mathbf{b}^{(i)}\}$ given the error $\boldsymbol{\epsilon}^{(i)}$ is
\begin{equation}
\label{ML=WLS}
 \mathcal{L}(r^{(i)}, t^{(i)},\mathbf{b}^{(i)}) = \frac{1}{\pi^N |\mathbf{R}|}\exp{\Big(-\boldsymbol{\epsilon}^{(i)H}\mathbf{R}^{-1}\boldsymbol{\epsilon}^{(i)}\Big)}\;,
 \end{equation} 
 where $^H$ denotes Hermitian transpose. 
Given these conditions, the ML estimation of the parameters $\{r^{(i)}, t^{(i)},\mathbf{b}^{(i)}\}$ is equivalent to a GLS problem \cite{ML74}:

\begin{equation}\label{maxmin}
\max_{r^{(i)}, t^{(i)},\mathbf{b}^{(i)}}\mathcal{L}(r^{(i)}, t^{(i)},\mathbf{b}^{(i)}) = \min_{r^{(i)}, t^{(i)},\mathbf{b}^{(i)}}\boldsymbol{\epsilon}^{(i)H}\mathbf{R}^{-1}\boldsymbol{\epsilon}^{(i)}\; ,
\end{equation}
where the constraints on $\mathbf{b}^{(i)}$ imposed in \eqref{minW} affect the solution, but do not alter the equality in \eqref{maxmin}. Therefore, if we assume that the residual vector is normally distributed, finding the weighting matrix $\mathbf{W}$ in \eqref{minW} is equivalent to estimating the covariance matrix of the residuals, i.e. we set $\mathbf{W}=\mathbf{{R}}^{-1}$. 
If the training dataset for determining $\mathbf{{R}}$ is small, it is typically preferable to constrain $\mathbf{W}$ to be diagonal in order to avoid overfitting. This is equivalent to the assumption of statistical independence between the residuals of each landmark.
\subsection{Empirical determination of the residual errors}
\label{ssec:test}
We need to estimate $\mathbf{R}$ based on a set of residual errors that are empirically determined from training images. We first simulate the search of the target landmarks using a detector $\mathcal{{T}}'$. We split our available training data set containing $2S$ images into two subsets of equal size. With the first subset we train $\mathcal{{T}}'$; with the second subset we measure the residual errors when employing $\mathcal{{T}}'$. For this, we simulate the detection of target landmarks, $\mathcal{{T}'}(\landmark^{(i-1)})$, for each of the $S$ images in the second subset, assuming that the true landmarks of each image are within the field of view of the detector $\mathcal{{T}'}$. This is achieved by setting $ \landmark^{(i-1)}=\landmark^\star_s + \boldsymbol{\delta}$, where $\boldsymbol{\delta} \in \mathbb{C}^N$ is a vector of translations small enough so that the true landmarks of the $s$th image, $\landmark_s^\star$, remain within the field of view of $\mathcal{{T}'}(\landmark^{(i-1)})$.  We employ the detector that searches for the best match of a gray-level profile as in \cite{Cootes96}; hence we determine $\boldsymbol{\delta}$ to place $\landmark^{(i-1)}$ on the line perpendicular to the true object contour, with translations small enough so that $\landmark_s^\star$ is sampled by $\mathcal{{T}'}(\landmark^{(i-1)})$. 

For each of the $S$ images, we perform $\mathcal{{T}'}(\landmark^\star_s + \boldsymbol{\delta}) = \hat{\landmark}_s$ and measure $\hat{\boldsymbol{\epsilon}}_{s} = \landmark^\star_s-\hat{\landmark}_s$, which is the residual error obtained from training sample $s$.
To determine the sample covariance matrix of the residuals we assemble the error matrix $ \hat{\mathbf{E}} = [\hat{\boldsymbol{\epsilon}}_{1},\dots,\hat{\boldsymbol{\epsilon}}_{S}] \in \mathbb{C}^{N\times S}$, which contains the residuals from all training samples. The sample covariance matrix is then
\begin{equation}
\hat{\mathbf{R}} = \frac{1}{S}\hat{\mathbf{E}}\hat{\mathbf{E}}^H .
\end{equation} 
We note that the so determined $\hat{\mathbf{R}}$ does not depend on the iteration $i$ of the ASM algorithm.

\subsection{Testing whether a target landmark is valid}
\label{ssec:FOV}

The sample covariance matrix of the residual errors $\hat{\mathbf{R}}$ is determined under the assumption that the true landmarks $\landmark^\star$ are within the field of view of the detectors $\mathcal{T}(\landmark^{(i-1)})$. If the true landmarks are not visible to the detector, the determined target landmarks are certainly incorrect. In such a case, the target landmarks are not valid.
In order to test whether a target landmark in $\tilde{\landmark}^{(i)}$ returned by the ASM algorithm is valid, we run a hypothesis test based on the metric that is used to determine the target landmark. 
In our case, this metric is a Mahalanobis distance $d^{(i)}_n$ that measures the distance between the observed and the modeled local appearance. Let us denote by $\mathbf{g}^{(i)}_n \in \mathbb{R}^{\ell\times1}$ the vector containing $\ell$ intensity values representing the appearance observed by detector $\mathcal{T}(\landmark^{(i-1)})$ to determine landmark $n$. The null hypothesis $\mathcal{H}_0$ is that the target landmark, which corresponds to the point that minimizes $d_n^{(i)}$, is valid. We assume $\mathbf{g}^{(i)}_n \sim \mathcal{N}(\boldsymbol{\mu}_{\mathbf{g}_n},\mathbf{S}_{\mathbf{g}_n})$. Thus, the Mahalanobis distance $d^{(i)}_n = (\mathbf{g}^{(i)}_n-\boldsymbol{\mu}_{\mathbf{g}_n})^{T}\mathbf{S}^{-1}_{\mathbf{g}_n}(\mathbf{g}^{(i)}_n-\boldsymbol{\mu}_{\mathbf{g}_n}) $ follows a chi-squared distribution with $\ell$ degrees of freedom under the null hypothesis:
\begin{align}
\mathcal{H}_0:  \mathsmaller{d^{(i)}_n  } \sim \mathlarger{\chi}^2_{\ell}
\end{align}
Therefore, for each iteration $i$ and each landmark $n$, we perform a chi-squared test of ${d^{(i)}_n}$: if the null hypothesis is not rejected, then we set an indicator variable $f^{(i)}_n = 1$; otherwise $f^{(i)}_n = 0$. We can fix a false alarm rate for this test based on the $\mathlarger{\chi}^2_{\ell}$ distribution. The idea of this approach is to work only with target landmarks whose observed gray-level profiles are close enough to the model profiles.

This test can be incorporated into the weighting matrix ${\mathbf{W}}$. For each iteration $i$, we define a diagonal matrix $\mathbf{F}^{(i)} = \text{diag}(f^{(i)}_1,$ $\dots,f^{(i)}_N)$. The weighting matrix ${\mathbf{W}}^{(i)}$ is then obtained as 
 \begin{align}
  {\mathbf{W}^{(i)}} &= \mathbf{F}^{(i)}\hat{\mathbf{R}}^{-1}\mathbf{F}^{(i)}.  
\end{align}

\section{Results and discussion}

\begin{figure}[h]
\centering
%
%
\definecolor{mycolor1}{rgb}{1.00000,0.60000,0.78431}%
\begin{tikzpicture}

\begin{axis}[%
width= \figurewidth,
height = \figurewidth,
at={(0in,0in)},
scale only axis,
point meta min=0,
point meta max=1,
axis on top,
xmin=0.5,
xmax=454.5,
tick align=outside,
y dir=reverse,
ymin=0.5,
ymax=454.5,
axis line style={draw=none},
legend style={legend cell align=left, align=left, draw=white!15!black},
ticks=none
]
\addplot [forget plot] graphics [xmin=0.5, xmax=454.5, ymin=0.5, ymax=454.5] {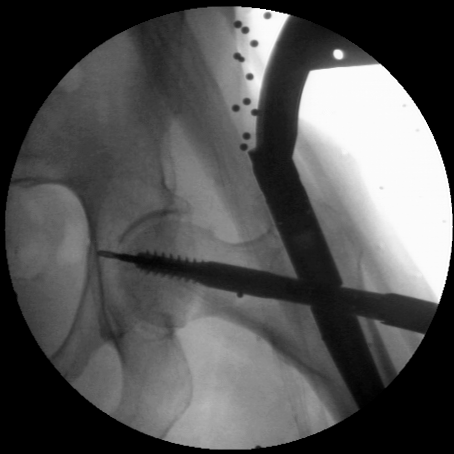};
\addplot [color=red, only marks, mark size=1.5pt, mark=*, mark options={solid, red}]
  table[row sep=crcr]{%
298	369\\
292	362\\
285	357\\
279	352\\
272	347\\
255	332\\
236	324\\
216	317\\
197	319\\
187	324\\
177	329\\
158	326\\
139	319\\
126	302\\
120	282\\
119	263\\
127	247\\
144	230\\
163	222\\
183	221\\
202	229\\
211	231\\
219	238\\
236	243\\
251	239\\
267	231\\
275	217\\
283	208\\
295	202\\
308	198\\
323	202\\
339	214\\
354	229\\
363	244\\
361	260\\
362	275\\
365	291\\
368	304\\
373	316\\
377	330\\
};
\addlegendentry{landmarks of the PDM}

\node[above left, align=right, font=\color{white}]
at (axis cs:310,371.883) { 1};
\node[above left, align=right, font=\color{white}]
at (axis cs:275,365) { 5};
\node[above left, align=right, font=\color{white}]
at (axis cs:220,317) { 8};
\node[above left, align=right, font=\color{white}]
at (axis cs:189.971,357) {11};
\node[above left, align=right, font=\color{white}]
at (axis cs:190,221) {20};
\node[above left, align=right, font=\color{white}]
at (axis cs:245,243) {24};
\node[above left, align=right, font=\color{white}]
at (axis cs:267,231) {26};
\node[above left, align=right, font=\color{white}]
at (axis cs:337,225) {32};
\node[above left, align=right, font=\color{white}]
at (axis cs:361,260) {35};
\node[above left, align=right, font=\color{white}]
at (axis cs:375.559,340.81) {40};

\end{axis}

\begin{axis}[%
width= \figurewidth,
height = \figurewidth,
at={(0in,0in)},
scale only axis,
xmin=0,
xmax=1,
ymin=0,
ymax=1,
axis line style={draw=none},
ticks=none,
axis x line*=bottom,
axis y line*=left
]
\node[below right, align=left, font=\bfseries\color{white}]
at (rel axis cs:0.471,0.4) {screw};
\node[below right, align=left, font=\bfseries\color{white}]
at (rel axis cs:0.63,0.497) {nail};
\end{axis}
\end{tikzpicture}%
\caption{Landmarks of a PDM (1 to 40) of the femur in a fluroscopic X-ray image in anterior-posterior orientation.}
\label{fig:PDM}
\end{figure}
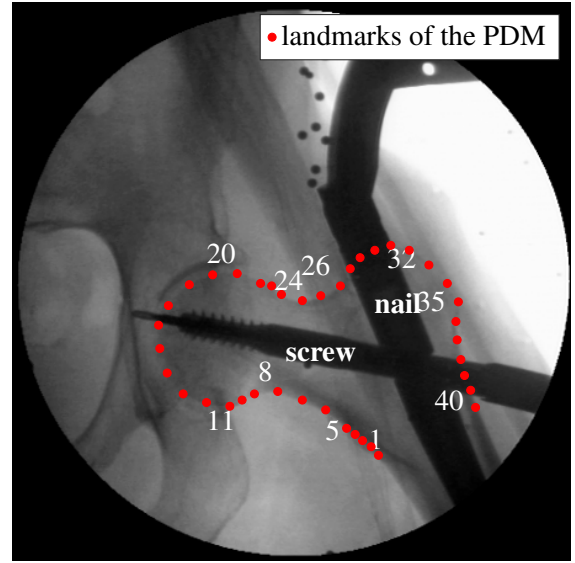

To evaluate the performance of our proposed method we choose a challenging segmentation task: the contour of the femur in fluoroscopic (low-dose) X-ray images. The quality of this image modality is low and thus the performance of the standard ASM algorithm suffers. Our database contains 350 gray-scale images that have been acquired during surgeries treating hip fractures in an approximate anterior-posterior orientation. These images show the upper part of the femur and part of the hip. They belong to different surgical interventions and C-arm devices. The image sizes range between 450x450 to 510x510 pixels. We have a ground truth consisting of a manually segmented femur contour and landmarks for every image. We show one of these images in Fig.~\ref{fig:PDM}. As these images come from surgeries treating hip fractures, they contain an intermedullary nail and a screw, which further complicate the segmentation task with occlusions. 
\subsection{Leave-one-out test}
\label{ssec:loo}
We evaluate the performance of our method using a leave-one-out test. For all the images in the database we ``leave one image out" to test and keep the remaining as training images. The process is repeated for every available image and the results are averaged over all test images. For each leave-one-out test iteration $t$, we obtain $\mathbf{\hat{P}}_{(t)}$, $\boldsymbol{\hat{\Lambda}}_{(t)}$ and detector $\mathcal{\hat{T}}_{(t)}$ from the training images. To perform the chi-squared test, for every landmark $n$ we also obtain $\hat{\boldsymbol{\mu}}_{\mathbf{g}_n(t)}$ and $\hat{\mathbf{S}}_{\mathbf{g}_n(t)}$. We determine the sample covariance of the residuals $\hat{\mathbf{R}}_{(t)}$ as described in Section \ref{ssec:test}.

We performed this leave-one-out test for all considered strategies. The same conditions (initial guess, shape parameters, number of ASM iterations...) were used for all, only changing the definition of the weighting matrix $\mathbf{W}$.  For each strategy, we ran 100 ASM iterations on each image. We then measured the squared distance of every resulting landmark to the ground truth contour. We show the root mean square error (RMSE) for each PDM landmark in Fig.~\ref{fig:results1}. 

\subsection{Evaluation of the proposed method}
\label{ssec:eval}
We implemented two different versions of our approach: one where we use the full matrix $\mathbf{W}$, and one where we restrict it to be diagonal. We compared these to the following strategies:
\begin{enumerate}
\item the standard ASM \cite{Cootes96} without weighting, i.e., $\mathbf{W} = \mathbf{I}$. 

\item $\mathbf{W}$ as a diagonal matrix with diagonal elements $1/\hat{d}_n$, as proposed by \cite{Zhao04}. 

\item $\mathbf{W}$ as a diagonal matrix with diagonal elements $1/(1+\text{trace}(\mathbf{S}_{\mathbf{g}_n}))$, as proposed by \cite{Yang14}. 
\end{enumerate}
As shown in Fig.~\ref{fig:results1}, our strategy outperforms all the other strategies on average, in particular when $\mathbf{W}$ is diagonal, i.e. assuming independent residuals (this could be due to the small sample size available). The improvement our strategy provides over other strategies is especially significant in the area of the femoral head (between landmark 11 and 20, with $\sim$30\% improvement achieved). This is the most challenging area of segmentation since there is overlap from other hip bones, and contour edges are particularly weak. Also, for the surgical procedure from which we obtained these images, this is an important region of interest \cite{Stryker11}.

As an example, we show in Fig.~\ref{fig:example} the landmarks found after 100 ASM iterations for our proposed weighting strategy with diagonal $\mathbf{W}$ compared to the standard ASM, for a particular image in the database. Our strategy follows the real contour much more closely.

\begin{figure} [h]
\centering
  \input{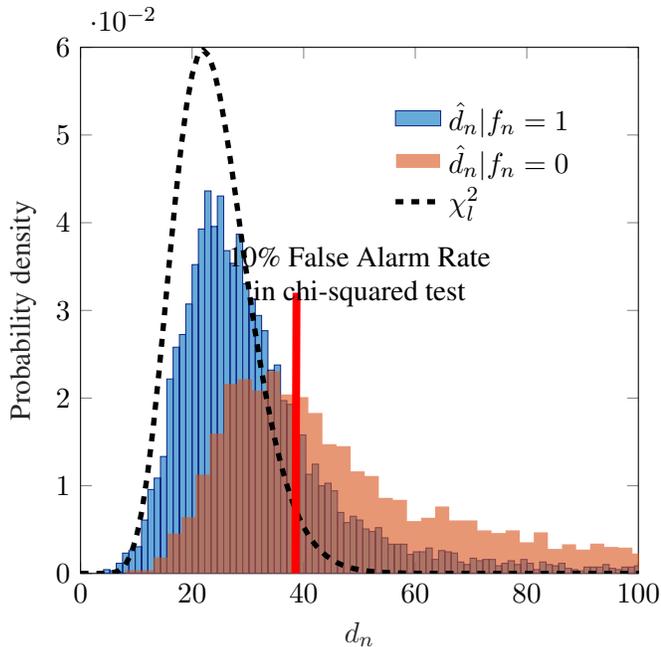}
\caption{Normalized histograms of the distances $\hat{d}_n | f_n=1$ and  $\hat{d}_n | f_n=0$ and the approximating chi-squared distribution for comparison.}
\label{fig:chi-square}
\end{figure}

We also further investigated the hypothesis test described in Section \ref{ssec:FOV}: We measured the norm $d_n$ after the simulation of several ASM iterations. Then we collected the observed $\hat{d}_n$ values either as $\hat{d}_n | f_n = 1$, if the true point was valid (null hypothesis), or $\hat{d}_n | f_n = 0$ otherwise. We computed normalized histograms of $\hat{d}_n$, as shown in Fig.~\ref{fig:chi-square}, to see how well a chi-squared distribution fits $\hat{d}_n | f_n = 1$. The approximation is fairly good, although it does not account for a strong tail in the histogram of $\hat{d}_n | f_n = 1$. We set a probability of false alarm, i.e. the probability of classifying a valid landmark as invalid,  of 10\% based on the approximating $\chi^2_\ell$ distribution. This covers $\sim$75\% of the values from $\hat{d}_n | f_n = 1$, but only $\sim$30\% from $\hat{d}_n | f_n = 0$. We suggest to set a relatively low false alarm threshold, which prevents the least-squares solution to be implausible. If too many landmarks are regarded as invalid and excluded, the problem may become underdetermined, and solutions may be unreliable.
\begin{figure}[h]
\centering
%
%
\definecolor{mycolor1}{rgb}{0.85098,0.32549,0.09804}%
\definecolor{mycolor2}{rgb}{0.00000,0.49804,0.00000}%
\definecolor{mycolor3}{rgb}{0.94118,0.94118,0.94118}%
\begin{tikzpicture}

\begin{axis}[%
clip mode=individual,
width= \figurewidth,
height = \figureheight,
at={(0in,0in)},
scale only axis,
xmin=1,
xmax=40.5,
xtick={ 1,  5,  9, 13, 17, 21, 25, 29, 33, 37, 40},
xlabel style={font=\color{white!15!black}},
xlabel={Landmark},
ymin=0,
ymax=24,
ylabel style={font=\color{white!15!black}, align=center},
ylabel={ RMSE (pixels)},
axis background/.style={fill=white},
xmajorgrids,
ymajorgrids,
legend pos=south east,
legend style={font =\scriptsize,anchor=south east, legend cell align=left, align=left, draw=white}
]
\addplot [color=black, line width=1.0pt, mark size=2.0pt, mark=+, mark options={solid, black}]
  table[row sep=crcr]{%
1	10.1213659955518\\
2	8.56160658912371\\
3	7.21123380444677\\
4	5.57593415641143\\
5	4.71888704241992\\
6	4.14643602286543\\
7	4.73463323106078\\
8	5.07065757303054\\
9	5.64423694117554\\
10	6.96386660378281\\
11	8.77714198302832\\
12	10.5572317837069\\
13	13.0419745158979\\
14	15.5611000200245\\
15	16.7917402344493\\
16	16.9279213229943\\
17	16.3619020577919\\
18	14.743360176996\\
19	13.4326420666079\\
20	11.8667816825308\\
21	10.4579430337936\\
22	9.02016371181615\\
23	8.10676803303141\\
24	10.3416478292068\\
25	14.492362170174\\
26	17.2197562378064\\
27	18.6405625142669\\
28	19.1927285727928\\
29	21.1082031302249\\
30	21.9444609230799\\
31	20.4815901359547\\
32	18.5246481773173\\
33	18.3622988650721\\
34	20.2401157900162\\
35	16.9448833998772\\
36	16.633569716272\\
37	17.1281521441923\\
38	17.7015976748228\\
39	18.8500311253871\\
40	21.9954879544715\\
};
\addlegendentry{standard ASM \cite{Cootes96}}

\addplot [color=mycolor1, dotted, line width=1.0pt, mark=+, mark options={solid, mycolor1}]
  table[row sep=crcr]{%
1	10.7410145972282\\
2	8.69262468153024\\
3	7.31881313891679\\
4	5.90147542707416\\
5	5.21074527566966\\
6	4.43053428162819\\
7	4.81131690031292\\
8	5.11616310681796\\
9	5.66378196950303\\
10	7.08075099224442\\
11	8.94824052859879\\
12	10.7910991293904\\
13	13.287286052742\\
14	15.7714519211103\\
15	16.9289422930622\\
16	16.8632912740502\\
17	15.9294416932155\\
18	14.2239200238227\\
19	12.9449850455023\\
20	12.2199107787333\\
21	9.78312109158547\\
22	8.63275971957286\\
23	7.80832960328108\\
24	10.3589297859559\\
25	14.5283528184427\\
26	17.4132036607899\\
27	18.8922503466724\\
28	18.790585569003\\
29	20.0135756223383\\
30	21.4089074688975\\
31	19.8563462237567\\
32	18.0499343414669\\
33	18.0135617405742\\
34	19.9734214825609\\
35	17.2631949402531\\
36	16.721175900986\\
37	17.1640062573657\\
38	17.5105674189165\\
39	19.0234697750996\\
40	21.8131849117443\\
};
\addlegendentry{$\mathbf{W}$ as in \cite{Yang14}}

\addplot [color=red, dashed, line width=1.0pt, mark size=2.0pt, mark=+, mark options={solid, red}]
  table[row sep=crcr]{%
1	9.62693541770105\\
2	7.6234059045376\\
3	6.45780183132953\\
4	5.29577303565652\\
5	4.64858288700487\\
6	3.89959102331051\\
7	4.31486941529534\\
8	5.24034709674218\\
9	5.83072603079544\\
10	6.74463572770658\\
11	8.68147598944868\\
12	10.8518397971285\\
13	13.8865709866912\\
14	16.6324373973916\\
15	17.7957924845334\\
16	17.6991726947201\\
17	17.1849164992905\\
18	15.2408609962397\\
19	13.7589362367704\\
20	12.4239336577542\\
21	10.4040407304844\\
22	8.93049767748946\\
23	8.10563991763079\\
24	10.1851939152677\\
25	14.5761794004596\\
26	17.4537458800639\\
27	19.344526182975\\
28	19.8064784249729\\
29	21.8534538525144\\
30	23.0606921124675\\
31	21.9164285615398\\
32	19.8933054546442\\
33	19.3745965906861\\
34	20.7512326596733\\
35	17.9820855348777\\
36	17.448695820452\\
37	17.8425576598587\\
38	17.9985103241366\\
39	19.1739658677377\\
40	21.6894143765489\\
};
\addlegendentry{$\mathbf{W}$ as in \cite{Zhao04}}

\addplot [color=mycolor2, dashed, line width=1.0pt, mark size=2pt, mark=square, mark options={solid, mycolor2}]
  table[row sep=crcr]{%
1	8.89947709690079\\
2	6.70195834423835\\
3	5.93344833827365\\
4	4.68976002152609\\
5	4.22789534872967\\
6	3.63135924862278\\
7	2.87225461291304\\
8	2.39985299044568\\
9	3.84471448406882\\
10	5.95569451856537\\
11	6.60812710438581\\
12	6.45303176141548\\
13	8.39695756346985\\
14	11.1403033091367\\
15	12.444060729586\\
16	12.1095439191196\\
17	10.9703267959085\\
18	9.57184737851709\\
19	8.7805232790102\\
20	8.74283700223024\\
21	8.77789310363898\\
22	9.04150951214921\\
23	8.3798910429573\\
24	9.00015497523727\\
25	11.5917775712025\\
26	14.6830665187531\\
27	14.1984100220384\\
28	14.3976243718917\\
29	16.1859366541332\\
30	17.5178883372057\\
31	17.2045626481434\\
32	16.0831135760711\\
33	16.3162608173876\\
34	18.674542272426\\
35	17.4070259246162\\
36	16.7064942344877\\
37	16.6701387618999\\
38	16.3525786122434\\
39	17.2138233608212\\
40	18.3430472309144\\
};
\addlegendentry{proposed $\mathbf{W}$ (diagonal)}

\addplot [color=blue, dashed, line width=1.0pt, mark size = 2pt, mark=asterisk, mark options={solid, blue}]
  table[row sep=crcr]{%
1	8.92108062555977\\
2	7.517127716277\\
3	6.76111786539701\\
4	5.75014103869493\\
5	4.82090819795894\\
6	3.88532141691978\\
7	3.11426545062093\\
8	3.04993886875501\\
9	4.21726260743157\\
10	6.47934504566548\\
11	7.81491469234889\\
12	8.02453215054288\\
13	10.0170299302837\\
14	12.5292706450067\\
15	13.9267494902347\\
16	13.6950912764347\\
17	12.437498788947\\
18	11.21132266372\\
19	10.7950639002554\\
20	10.3812261040077\\
21	10.5203414515308\\
22	9.84805253430509\\
23	9.40181840013878\\
24	11.349673882081\\
25	14.5474611331997\\
26	16.6850659106416\\
27	14.863048594461\\
28	14.2290967162266\\
29	16.3338477687727\\
30	18.5870805616345\\
31	18.9095813049346\\
32	18.810045955616\\
33	19.8851215295237\\
34	22.4683694850146\\
35	20.6769285240358\\
36	18.8283941046321\\
37	17.8282724342047\\
38	17.1646750630746\\
39	17.4522648941735\\
40	19.704736393384\\
};
\addlegendentry{proposed $\mathbf{W}$ (full)}

%
\draw[draw=none, opacity=0.7,fill=mycolor3] (axis cs:25.5,0) rectangle (axis cs:31.5,24);
\draw[draw=none, opacity=0.7,fill=mycolor3] (axis cs:37.5,0) rectangle (axis cs:40.5,24);
\node[below right, align=left, draw=none, rotate=90]
at (axis cs:38,10) {screw};
\node[below right, align=left, draw=none, rotate=90]
at (axis cs:26.5,10) {nail};
\end{axis}
\end{tikzpicture}%
\caption{RMSE of our proposed strategy compared with \cite{Yang14}, \cite{Zhao04} and \cite{Cootes96}. We mask in gray the areas where the surgical implants (nail and screw) most likely occlude the bone contour.}
\label{fig:results1}
\end{figure}
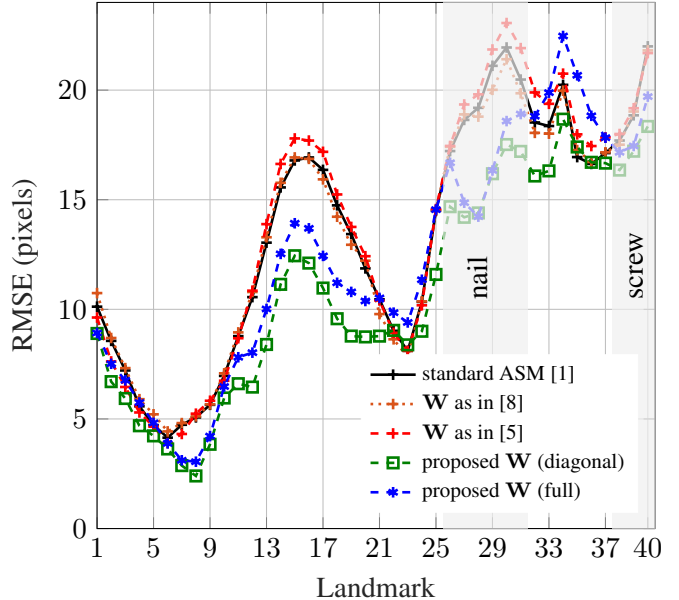
\begin{figure}[h!]
\centering
   \input{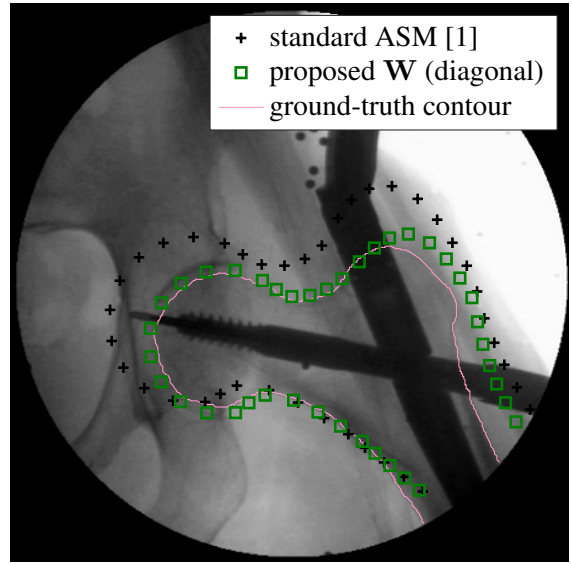}
\caption{One example of a segmented femur, comparing our technique with the standard ASM algorithm (without weighting).}
\label{fig:example}
\end{figure}
\section{Conclusions}
\label{sec:concl}
We proposed a GLS instead of an ordinary least-squares fit as a straightforward and simple strategy to add robustness to the ASM algorithm. The weights in the GLS fit should not be selected only assuming heuristic confidence metrics of the landmarks, since these do not generalize and are prone to over-fitting. We proposed to choose the weights based on empirically determined errors of the least-squares fit, where the weighting matrix $\mathbf{W}$ is determined as the inverse of the sample covariance of the residuals. This can be interpreted as a Maximum Likelihood solution of the least-squares problem. Additionally we used a chi-squared test to identify target landmarks that are likely to be incorrect and thus exclude them from the fit. Therefore, we combine prior knowledge about the performance of the target detectors, $\hat{\mathbf{R}}$, with additional information about the unreachable landmarks (either occluded or out of alignment), $\mathbf{F}$.
We tested the strategy in fluoroscopic X-ray images of the femur taken in actual surgeries. We showed that our strategy outperforms the standard ASM as well as other weighting strategies. While our approach was based on a particular metric for identifying target landmarks (the Mahalanobis distance), our idea should generalize to other metrics as well.

\bibliographystyle{IEEEbib}
\bibliography{references}

\end{document}